\documentclass[11pt,twocolumn]{article}

\usepackage[utf8]{inputenc}
\usepackage[T1]{fontenc}
\usepackage{times}
\usepackage[margin=0.75in,columnsep=0.25in]{geometry}
\usepackage{microtype}
\usepackage{amsmath,amssymb,amsthm,amsfonts}
\usepackage{mathtools}
\usepackage{graphicx}
\usepackage{booktabs}
\usepackage{multirow}
\usepackage{array}
\usepackage{tabularx}
\usepackage{enumitem}
\setlist{nosep,leftmargin=*}
\usepackage[dvipsnames]{xcolor}
\usepackage{tikz}
\usepackage{pgfplots}
\pgfplotsset{compat=1.18}
\usetikzlibrary{arrows.meta,positioning,shapes.geometric,fit,backgrounds,calc,decorations.pathreplacing}
\usepackage{caption}
\usepackage[hidelinks]{hyperref}
\usepackage{cleveref}
\usepackage[ruled,vlined,linesnumbered]{algorithm2e}
\usepackage{titlesec}
\titlespacing*{\section}{0pt}{2.0ex plus 0.5ex minus .2ex}{1.0ex plus .2ex}
\titlespacing*{\subsection}{0pt}{1.5ex plus 0.3ex minus .2ex}{0.8ex plus .2ex}
\titlespacing*{\subsubsection}{0pt}{1.0ex plus 0.2ex minus .1ex}{0.5ex plus .1ex}

\newcommand{\sys}{\textsc{Rhetor}}

\definecolor{rhblue}{RGB}{30,80,160}
\definecolor{rhgreen}{RGB}{40,140,90}
\definecolor{rhorange}{RGB}{210,120,40}
\definecolor{rhpurple}{RGB}{120,60,160}
\definecolor{rhred}{RGB}{200,60,60}

\newtheorem{definition}{Definition}

\newtheorem{invariant}{Invariant}

\title{\Large \textbf{Rehearsed Multi-Agent Live Product Demonstrations\\with Real-Time Voice Question Answering}}

\author{%
Rahul Khedar$^{*}$, Mayank Malhotra$^{*}$\\
Avinash Karn, Mouli V, Prakhar Mehrotra\\[8pt]
\textit{PayPal AI}\\[4pt]
}
\date{}

\begin{document}
\maketitle

{\noindent\Large\bfseries Abstract}
\\
\\
\noindent
Live product demonstrations are a recurring, high-cost activity in software organizations: a human presenter must select features, dispatch the corresponding interactions on a running application, narrate them coherently, and answer questions in real time. Existing automation addresses only fragments---generalist browser agents target instruction-conditioned task completion, and demo-video tools produce fixed MP4 artifacts that cannot be questioned and silently break under interface drift. We propose \sys{}, a multi-agent system that takes a running web application and its source-code repository as input and produces a rehearsed live demonstration with segment-synchronized narration and real-time voice question answering. The architectural contributions are a cross-modal feature representation that merges UI exploration with source-code analysis into features tagged with discrete focus tiers, a grounded scripter constrained to UI elements observed during exploration and dispatched through multi-strategy semantic locators, a pre-presentation rehearsal loop with explicit convergence and graceful degradation to narration-only segments, and a runtime synchronization invariant that ties each browser action to the audio-end event of its narration segment. Across six pipeline sessions on four deployed applications---including the public-domain whiteboard application Excalidraw---the rehearser's internal locator-firing rate $\bar\sigma$ spans $0.31$--$1.00$ over $147$ scripted actions; on the substantial workload (53 actions, full tier differentiation), $\bar\sigma \approx 0.92$, and on the public-domain reference point the locator-repair step drives convergence to $\bar\sigma = 1.00$ at iteration $2$. We additionally define a benchmark protocol of ten metrics across six application categories that would establish, beyond the case study, whether each design choice contributes positively.

\section{Introduction}
\label{sec:intro}

Live product demonstrations are a recurring activity in software organizations across sales engineering, developer relations, customer success, and engineering onboarding. The activity is structurally constrained: in a bounded time window, a human presenter selects a subset of features from an application, executes the corresponding interactions on the running product, narrates the interactions coherently, and responds to audience questions whose subject matter spans both user-facing behavior and architectural detail. The labor cost is substantial, the activity repeats indefinitely across audience rotations, and the produced artifact---an instance of internalized state in the recipient---is invalidated by the next interface change.

Two classes of automation address subsets of this problem. Generalist browser agents~\cite{deng2023mind2web,zhou2023webarena,koh2024visualwebarena,he2024webvoyager,molmoweb2026,xue2025illusion} target instruction-conditioned task completion: given a natural-language goal, the agent attempts to complete the task on a real website, with success measured by outcome. The demonstration problem differs in two respects: the system must \emph{select} which features to demonstrate rather than be told, and it must verbalize what it is doing rather than only execute it. A second class of tools~\cite{demopilot,look,neurascreen,demodsl} composes a vision LLM, a browser-automation library, and a text-to-speech model to render narrated MP4 videos from a target URL. The video format produces a fixed artifact: it cannot accept runtime questions, cannot adapt to user pacing, and fails silently when the underlying interface has drifted from its state at render time. Vardanyan~\cite{vardanyan2025browser}, reporting on a year of production browser-agent operation, argues that the dominant factor governing real deployment is architectural rather than model-capability-driven; we adopt this observation as a working premise.

We propose \sys{}, a multi-agent system for rehearsed product demonstrations of web applications under live runtime conditions. Given as input a running application and its source-code repository, \sys{} executes a five-phase pipeline---explore, code-read, understand, script, rehearse---and produces, as output, a rehearsed demonstration that is served to a client browser through a same-origin reverse proxy embedded in the user's browser, with audio narration synchronized to browser actions through a per-segment handshake and user speech bridged to a real-time speech-to-speech endpoint grounded in a generated knowledge document.

\paragraph{Contributions.} (i) We formalize a cross-modal feature representation in which UI exploration and source-code analysis are jointly merged into features tagged with discrete focus tiers $\mathcal{T} = \{\textsc{hero}, \textsc{supp}, \textsc{mention}\}$ and a continuous demo priority $\rho \in [1,10]$; the focus tier operationalizes a narrative attention budget that downstream phases enforce. (ii) We constrain the scripter to ground every action in an element observed during exploration; each action is dispatched through an ordered locator tuple $\mathcal{L} = (\ell_1, \ldots, \ell_6)$ over the strategies \emph{role+name}, \emph{text}, \emph{label}, \emph{placeholder}, \emph{test-id}, and \emph{CSS}, tried in fixed priority order. (iii) We introduce a rehearse-then-present loop with explicit convergence and degradation: the script is executed in a real headless browser, an LLM proposes locator alternatives for failed actions, locator lists update by left-prepend $L_{i+1}(a) \leftarrow R_i(a) \oplus L_i(a)$, iteration continues while $\sigma_i < \tau$ up to a fixed cap $I_{\max}$, and any action that the loop cannot validate is converted to a narration-only segment rather than removed, preserving termination of the runtime trace. (iv) We specify a runtime synchronization invariant: the unit of authoring is the narration segment $\sigma = \langle \mathit{text}, \mathit{action}? \rangle$, and the runtime guarantees $t^{(\sigma)}_{\mathrm{action}} = t^{(\sigma)}_{\mathrm{audio\_end}}$ via a server--client handshake, eliminating the word-offset drift characteristic of TTS-aligned automation under variable provider latency.

Beyond the four core contributions, the paper specifies a real-time voice question-answering path that bridges browser PCM to a server-VAD speech-to-speech endpoint with the generated knowledge document injected as session instructions, and proposes a benchmark protocol of ten metrics across six application categories. A preliminary case study on four deployed applications, exercising multiple rehearsal regimes, is reported in \Cref{sec:case}.

The remainder of the paper is organized as follows. \Cref{sec:related} reviews prior work. \Cref{sec:prelim} fixes notation. \Crefrange{sec:phase1}{sec:runtime} describe the five phases and the runtime. \Cref{sec:impl} reports implementation details. \Cref{sec:eval} defines the benchmark protocol. \Cref{sec:case} reports the case study. \Cref{sec:limits,sec:disc} discuss limitations and design implications.

\section{Related Work}
\label{sec:related}

\paragraph{Generalist browser agents.}
WebArena~\cite{zhou2023webarena} provides reproducible Docker-sandboxed web applications with outcome-based evaluation; VisualWebArena~\cite{koh2024visualwebarena} adds a vision dimension. Mind2Web~\cite{deng2023mind2web} contributes 2{,}350 tasks across 137 websites with cross-task, cross-website, and cross-domain splits. WebVoyager~\cite{he2024webvoyager} reports an end-to-end multimodal agent and a benchmark of the same name. Recent work~\cite{xue2025illusion} re-evaluates state-of-the-art agents on Online-Mind2Web and reports substantially lower numbers than headline scores suggest. The open-weights MolmoWeb~\cite{molmoweb2026} releases a vision web agent with full training and evaluation tooling. These systems pursue task-completion under instruction; \textsc{Rhetor} pursues narrative selection and presentation. Action selection occurs at \emph{author time} (rehearsal) rather than \emph{run time}, so the live runtime consumes a validated trace rather than acting under uncertainty.

\paragraph{Production browser-agent reports.}
Vardanyan~\cite{vardanyan2025browser} reports a year of production operation of a browser agent and concludes that hybrid accessibility-tree-plus-vision context, specialization over general autonomy, and programmatic safety boundaries determine reliability more than LLM scale does. We adopt this stance and contribute the additional pattern of \emph{rehearse-then-present}: relocating the agent's failure surface to a controlled offline phase with a defined degradation rule.

\paragraph{Automated demo videos.}
A recent class of tools~\cite{demopilot,look,neurascreen,demodsl} maps a URL or YAML specification to a narrated MP4 via the pattern: capture page $\to$ vision-LLM analysis $\to$ script $\to$ record browser $\to$ TTS $\to$ ffmpeg merge. \textsc{Rhetor} differs in three respects: the output is a live, interactive demonstration in the target application served via a same-origin reverse proxy rather than a pre-rendered video; planning consumes both UI and source-code signals, so architectural content (integrations, scalability, data model) becomes addressable demo material; and a pre-presentation rehearsal phase converts a best-effort script into a validated one before deployment in a deployed setting.

\paragraph{Speech-to-speech runtimes.}
Real-time audio APIs~\cite{openai2024realtime} expose server-side voice activity detection and direct speech-to-speech generation over WebSocket, eliminating the transcribe--think--synthesize round-trip. We use such an endpoint as the carrier for live voice question answering, with a generated knowledge document injected as session instructions to ground responses in the demonstrated application.

\paragraph{State-grounded multi-agent generation.}
Khedar et al.~\cite{khedar2025stategen} introduce an authoritative state object that constrains LLM outputs across a multi-role generation loop in the setting of synthetic data generation. We adopt the same principle in a different setting: a non-LLM data structure---the merged \texttt{SiteMap} and \texttt{CodeAnalysis}---constrains LLM outputs across the pipeline phases.

\section{Preliminaries and Notation}
\label{sec:prelim}

We fix notation used throughout the paper.

\paragraph{Application surface.}
Let $G = (V, E)$ denote the navigation graph of the target application, where $V$ is the set of pages discovered by exploration and $E \subseteq V \times V$ is the set of inter-page links. Each page $v \in V$ carries a tuple $\langle \mathit{path}, \mathit{title}, \mathit{type}, \mathit{summary}, \mathcal{E}_v, \mathcal{F}_v, \mathcal{M}_v \rangle$ where $\mathcal{E}_v$ is the set of observed interactive elements, $\mathcal{F}_v$ the set of forms, and $\mathcal{M}_v$ the set of modals. An element $e \in \mathcal{E}_v$ has the structured signature $\langle \mathit{role}, \mathit{text}, \mathit{aria}, \mathit{label}, \mathit{placeholder}, \mathit{testid}, \mathit{bbox} \rangle$.

\paragraph{Code surface.}
Let $\mathcal{R}$ denote the set of routes extracted from the repository, $\mathcal{D}$ the set of data models, and $\mathcal{F}_\mathrm{rw}$ the inferred web framework. Together, the \emph{CodeAnalysis} object is $\mathcal{C} = \langle \mathcal{R}, \mathcal{D}, \mathcal{F}_\mathrm{rw}, \alpha \rangle$ with $\alpha$ a free-text architecture summary.

\paragraph{Features and tiers.}
A feature $f \in \mathcal{F}$ is a tuple $\langle \mathit{name}, P_f, K_f, \tau_f, \rho_f, \theta_f \rangle$ where $P_f \subseteq V$ are the pages over which $f$ spans, $K_f \subseteq \bigcup_{v \in P_f} \mathcal{E}_v$ is the set of key elements, $\tau_f \in \mathcal{T} = \{\textsc{hero}, \textsc{supp}, \textsc{mention}\}$ is the focus tier, $\rho_f \in [1,10]$ the demo priority, and $\theta_f$ an optional architecture-level note.

\paragraph{Locators.}
A locator strategy is a function $\ell : (\mathit{page}, \mathit{element}) \mapsto \{\mathit{handle}, \emptyset\}$ that returns either a Playwright element handle or fails. We fix a priority-ordered tuple of strategies $\mathcal{L} = (\ell_1, \ldots, \ell_6)$ summarized in \Cref{tab:locators}. For an action target $t$, the \emph{winning locator} is
\begin{equation}
\ell^*(t) \;=\; \ell_{i^*}(t), \quad i^* = \min\{ i : \ell_i(t) \neq \emptyset \},
\end{equation}
or undefined if no strategy matches.

\begin{table}[h]
\centering
\small
\setlength{\tabcolsep}{5pt}
\begin{tabular}{@{}clp{3.4cm}@{}}
\toprule
$i$ & Strategy $\ell_i$ & Selector pattern \\
\midrule
1 & role+name      & \texttt{getByRole(role,\{name\})} \\
2 & text           & visible text content \\
3 & label          & associated \texttt{<label>} \\
4 & placeholder    & input \texttt{placeholder} attr.\ \\
5 & test-id        & \texttt{[data-testid="..."]} \\
6 & CSS            & raw CSS selector (fallback) \\
\bottomrule
\end{tabular}
\caption{The locator strategy tuple $\mathcal{L}$. Strategies are tried in priority order; the first match becomes $\ell^\star(t)$. The ordering favors semantic-role and text-based strategies, which are robust to incidental DOM changes, over CSS, which couples to internal structure.}
\label{tab:locators}
\end{table}

\paragraph{Script.}
A demo script $S$ is a sequence of scenes $S = (s_1, \ldots, s_n)$. Each scene $s$ comprises an act $\mathit{act}(s) \in \{\textsc{hook}, \textsc{journey}, \textsc{hood}, \textsc{close}\}$, an entry path, and a sequence of \emph{narration segments} $\Sigma(s) = (\sigma_1^s, \ldots, \sigma_{m_s}^s)$. Each segment is the pair $\sigma = \langle \mathit{text}, \mathit{action}? \rangle$, where the action is optional (allowing pure narration).

\paragraph{Validated actions.}
After rehearsal, each action $a$ is annotated with a status $\phi(a) \in \{\textsc{verified}, \textsc{failed}, \textsc{narration\_only}\}$, the winning locator $\ell^*(a)$, and a duration. A scene's success rate is
\begin{equation}
\sigma(s) \;=\; \frac{|\{ a \in \mathcal{A}(s) : \phi(a) = \textsc{verified} \}|}{|\mathcal{A}(s)|},
\end{equation}
where $\mathcal{A}(s) = \{a : \exists \sigma \in \Sigma(s),\, \mathit{action}(\sigma) = a\}$. The overall success rate of the script is the size-weighted mean $\bar{\sigma} = \sum_s |\mathcal{A}(s)| \sigma(s) / \sum_s |\mathcal{A}(s)|$.

\paragraph{Objective.}
The system input is the pair $(U_{\mathrm{app}}, U_{\mathrm{repo}})$ of an application URL and a repository URL. The system output is a tuple $\langle S^*, \Phi^*, K^*, U_{\mathrm{app}} \rangle$ where $S^*$ is a rehearsed script, $\Phi^*$ the validated-action annotation, $K^*$ a Markdown knowledge document, and $U_{\mathrm{app}}$ the same application URL, served live to a client browser with both text and voice question-answering grounded in $K^*$.

\begin{figure*}[t]
\centering
\begin{tikzpicture}[
  font=\footnotesize,
  every node/.style={align=center},
  phase/.style={rectangle,rounded corners=3pt,draw=black!50,thick,minimum width=2.0cm,minimum height=1.0cm,fill=white,inner sep=4pt},
  arr/.style={-{Latex[length=2mm]},thick,black!70},
  edgelbl/.style={font=\scriptsize\itshape,fill=white,inner sep=1pt,text=black!70},
]
\node[phase,fill=rhblue!12] (explore) {Phase 1a\\\textbf{Explore}};
\node[phase,fill=rhblue!12,below=0.55cm of explore] (code) {Phase 1b\\\textbf{Code Read}};

\node[phase,fill=rhgreen!12,right=1.6cm of $(explore)!0.5!(code)$] (understand) {Phase 2\\\textbf{Understand}};
\node[phase,fill=rhorange!12,right=0.95cm of understand] (script) {Phase 3\\\textbf{Script}};
\node[phase,fill=rhpurple!12,right=0.95cm of script] (rehearse) {Phase 4\\\textbf{Rehearse}};
\node[phase,fill=rhred!12,right=0.95cm of rehearse] (present) {Phase 5\\\textbf{Present}};

\draw[arr] (explore.east) -- node[edgelbl,pos=0.55]{$G$} (understand.west);
\draw[arr] (code.east) -- node[edgelbl,pos=0.55]{$\mathcal{C}$} (understand.west);
\draw[arr] (understand) -- node[edgelbl]{$\mathcal{F},K$} (script);
\draw[arr] (script) -- node[edgelbl]{$S$} (rehearse);
\draw[arr] (rehearse) -- node[edgelbl]{$\langle S^*,\Phi^*\rangle$} (present);

\node[draw=black!40,dashed,rounded corners=3pt,fit=(explore)(code),inner sep=6pt,label={[font=\scriptsize\itshape]above:parallel ($T_1=\max(T_{1a},T_{1b})$)}] (par) {};
\end{tikzpicture}
\caption{The five-phase \textsc{Rhetor} pipeline. Phases 1a and 1b execute concurrently and produce the navigation graph $G$ and code analysis $\mathcal{C}$ consumed by Phase 2. Edge labels denote the typed artifact passed between phases; full definitions are in \Cref{sec:prelim}.}
\label{fig:pipeline}
\end{figure*}
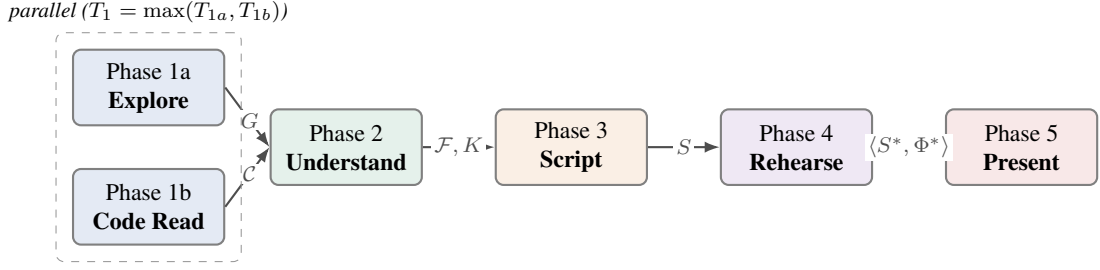

\section{Phase 1: Exploration and Code Reading}
\label{sec:phase1}

Phases 1a and 1b run concurrently. Phase 1a constructs $G$ via a bounded breadth-first crawl of the running application; Phase 1b constructs $\mathcal{C}$ from the source repository. The two phases share no state and their wall-clock cost differs in regime, so end-to-end Phase 1 latency is
\begin{equation}
T_1 \;=\; \max\bigl(T_{1a}, T_{1b}\bigr) \;\le\; T_{1a} + T_{1b}.
\end{equation}

\subsection{UI exploration}
\label{sec:explore}

The crawl is bounded by $|V| \le \kappa_p$ and $\mathrm{depth}(v) \le \kappa_d$ for $\kappa_p = 30$ and $\kappa_d = 4$ by default. Page perception is hybrid: deterministic JavaScript executed in the page context extracts $\mathcal{E}_v$ over a selector union (\texttt{a, button, [role], [data-testid]}, headings, form controls, labeled inputs), restricted to the visible viewport; a multimodal LLM call then summarizes the page from its 1280$\times$720 PNG screenshot, producing the structured triple $\langle \mathit{type}, \mathit{summary}, \mathit{name} \rangle$. Page-readiness combines Playwright's network-idle signal with a JavaScript spinner detector polling at 250\,ms intervals.

Three behaviors extend the basic crawl. \emph{Modal probing} clicks up to five candidate triggers per page in fresh contexts to discover dialogs absent from the static crawl. \emph{Vision-assisted login} resolves authentication walls in the absence of configured credentials by issuing a vision LLM call that identifies a public-area click target, after which a JavaScript tree-walker performs the click; the same component identifies SSO entry points (OAuth, SAML, Auth0, Okta) and persona-card login flows characteristic of enterprise SaaS deployments, which are not handled by a generic password-field heuristic. \emph{App-type classification} labels the application as SPA, MPA, or hybrid via URL-pattern heuristics; the runtime reverse proxy uses this label.

\subsection{Code reading}
\label{sec:code}

When a GitHub access token is configured, the system fetches the repository through the REST API: a single recursive tree request lists all blobs, files are filtered by extension and a directory deny list, and contents are fetched in parallel by a worker pool of size 10. No clone is performed and no source is written to disk. Without a token, a shallow clone (\texttt{--depth 50}) is used as fallback.

The fetched files are partitioned into batches by a path-aware batcher with byte budget $B \approx 60$\,KB, cohorting files by directory. Each batch is analyzed by an LLM under a structured-output prompt and the per-batch results are merged with route deduplication keyed by \texttt{method:path}. Two additional one-shot calls produce the framework label $\mathcal{F}_\mathrm{rw}$ from the directory tree and the architecture summary $\alpha$.

\section{Phase 2: Cross-Modal Understanding}
\label{sec:phase2}

The merge $\Psi : (G, \mathcal{C}) \mapsto (\mathcal{F}, K)$ is a single LLM call under a structured-output prompt; the prompt receives a serialized form of the navigation graph (pages with summaries and key elements) and the code analysis (routes, models, framework). Its outputs are the feature set $\mathcal{F}$ (with $\tau_f, \rho_f, K_f, P_f, \theta_f$ populated for each $f$) and the prompt for a second call that produces the Markdown knowledge document $K$ with prescribed sections: \emph{Product Overview}, \emph{Features}, \emph{Architecture}, \emph{Common Q\&A}, \emph{Talking Points}. The \emph{Common Q\&A} section is generated anticipatorily: the model is prompted to enumerate ten to fifteen plausible audience questions and to produce reference answers grounded in $\mathcal{F}$ and $\mathcal{C}$ at generation time, rather than at runtime. Pre-computation of the question set is a structural advantage relative to a human presenter, who typically prepares answers reactively.

\paragraph{Focus tier as attention budget.}
Let $w(\sigma) \in \mathbb{N}$ denote the word count of segment $\sigma$. The system enforces tier-conditional bounds at narration time:
\begin{equation}
w(\sigma) \in
\begin{cases}
[100, 140] & \text{if } \tau_{f(\sigma)} = \textsc{hero} \\
[80, 110] & \text{if } \tau_{f(\sigma)} = \textsc{supp} \\
[40, 65] & \text{if } \tau_{f(\sigma)} = \textsc{mention}
\end{cases}
\end{equation}
where $f(\sigma)$ is the feature associated with $\sigma$ via its scene. The tier is a coarse, prompt-enforceable signal that prevents an LLM scripter from producing flat, equally-weighted walkthroughs.

\paragraph{Knowledge index.}
$K$ is partitioned into sections $\{c_1, \ldots, c_N\}$ by splitting on level-2 headers. Each section is embedded once via an OpenAI-compatible embedding endpoint. Retrieval at query time uses cosine similarity
\begin{equation}
s(q, c) \;=\; \frac{\mathbf{e}_q \cdot \mathbf{e}_c}{\|\mathbf{e}_q\|_2\,\|\mathbf{e}_c\|_2},
\end{equation}
with a keyword-overlap fallback when an embedding endpoint is unavailable. The retrieval context for query $q$ is
\begin{equation}
\mathcal{R}(q) \;=\; \arg\!\!\!\max\nolimits_{C \subseteq \mathrm{Top}_k(q),\; \|C\| \le B_q} \sum_{c \in C} s(q, c),
\end{equation}
where $B_q$ is a character budget (default 4000) and $\|C\|$ is the total character length of the selected sections.

\section{Phase 3: Grounded Demo Scripting}
\label{sec:phase3}

The scripter produces $S$ under the structural constraint
\begin{equation}
\mathit{act}(s_1) = \textsc{hook},\;\; \mathit{act}(s_n) = \textsc{close},
\end{equation}
with $\mathit{act}$ taking values \textsc{journey} for $|\mathit{act}|=3$--$6$ middle scenes and \textsc{hood} for $1$--$2$ technical scenes drawn from $\theta_f$ and $\alpha$. The grounding constraint is the predicate
\begin{equation}
\forall \sigma \in \bigcup_s \Sigma(s),\;\; \mathit{action}(\sigma) \neq \emptyset \;\Rightarrow\; \mathit{target}(\sigma) \in \bigcup_v \mathcal{E}_v,
\end{equation}
i.e., every action targets an element observed during exploration. The constraint is enforced both in the prompt (the LLM is given the element list) and in post-processing: an unmatched target is downgraded to a single text-based locator and flagged for rehearsal scrutiny.

The action vocabulary is
$$
\mathcal{V} = \{\texttt{navigate}, \texttt{click}, \texttt{fill}, \texttt{hover},$$
$$\texttt{scroll}, \texttt{wait}, \texttt{highlight}\}.$$
\texttt{highlight} draws a temporary outline and scrolls the target into view; the rehearser uses it for narration alignment and the runtime uses it as a presenter pointer.

\paragraph{Realistic data.}
For \texttt{fill} actions, the scripter invokes a small structured-output LLM call that generates plausible demo values (names, dates, currency, descriptive text) rather than placeholder strings; this is qualitatively observable in user perception even when not measurable in success-rate metrics.

\section{Phase 4: Rehearsal with LLM Locator Repair}
\label{sec:phase4}

\paragraph{Iteration.}
At iteration $i$, every action $a$ in $S$ is dispatched in a real headless Chromium browser; the executor tries strategies in $\mathcal{L}$ in priority order and returns $\ell^*(a)$ or $\emptyset$. Each $a$ is annotated with $\phi_i(a) \in \{\textsc{verified}, \textsc{failed}\}$, the winning locator, the post-action bounding box, the duration, and an error message if any. Let $\sigma_i = \bar\sigma$ be the overall success rate after iteration $i$.

\paragraph{Repair.}
For each $a$ with $\phi_i(a) = \textsc{failed}$, an LLM call under a structured-output prompt proposes new locator strategies $R_i(a)$. The action's locator list is updated by left-prepend
\begin{equation}
L_{i+1}(a) \;\leftarrow\; R_i(a) \oplus L_i(a),
\end{equation}
which preserves the original strategies as fallbacks. Iteration continues while
\begin{equation}
\sigma_i < \tau \;\wedge\; i < I_{\max},
\end{equation}
with default convergence threshold $\tau = 0.95$ and iteration cap $I_{\max} = 3$.

\paragraph{Degradation.}
On termination at $i^* = I_{\max}$ with $\sigma_{i^*} < \tau$, the residual failed actions are not removed from $S$. Instead the rule
\begin{equation}
\phi(a) := \textsc{narration\_only} \quad \text{for all } a \text{ with } \phi_{i^*}(a) = \textsc{failed}
\end{equation}
is applied, and the script is marked $\mathit{ready}$ when $\sigma_{i^*} \ge 0.7$. At runtime, segments whose action is \textsc{narration\_only} play the spoken text and skip the browser interaction. This ensures completion of the live presentation under any failure mode reachable via the rehearsal loop.

\paragraph{Action success at the script level.}
We use action-success-at-1, mirroring the locator priority order:
\begin{equation}
\mathrm{Succ@1}(S) \;=\; \frac{1}{|\mathcal{A}|}\sum_{a \in \mathcal{A}} \mathbf{1}[\phi(a) = \textsc{verified}].
\end{equation}

\Cref{alg:rehearse} summarizes the full procedure.

\begin{algorithm}[t]
\SetAlgoLined
\DontPrintSemicolon
\KwIn{script $S$, locator strategies $\mathcal{L}$, threshold $\tau$, cap $I_{\max}$}
\KwOut{rehearsed script $S^\star$ with status map $\Phi^\star$}
\For{$i \leftarrow 1$ \KwTo $I_{\max}$}{
  \For{$a \in \mathcal{A}(S)$}{
    $h \leftarrow \ell^\star(a)$ \tcp*{first matching locator in $\mathcal{L}$}
    \eIf{$h \neq \emptyset$}{
      execute $a$ on $h$\;
      $\phi_i(a) \leftarrow \textsc{verified}$\;
    }{
      $\phi_i(a) \leftarrow \textsc{failed}$\;
    }
  }
  $\sigma_i \leftarrow \mathrm{Succ@1}(S \mid \phi_i)$\;
  \If{$\sigma_i \geq \tau$}{
    \Return $\langle S, \phi_i \rangle$ with $\mathit{ready} \leftarrow \mathbf{T}$\;
  }
  \ForEach{$a$ with $\phi_i(a) = \textsc{failed}$}{
    $R_i(a) \leftarrow \textsc{LLM-Repair}(a, \mathit{err}(a))$\;
    $L_{i+1}(a) \leftarrow R_i(a) \oplus L_i(a)$ \tcp*{prepend}
  }
}
\ForEach{$a$ with $\phi_{I_{\max}}(a) = \textsc{failed}$}{
  $\phi(a) \leftarrow \textsc{narration\_only}$ \tcp*{degrade}
}
$\mathit{ready} \leftarrow [\sigma_{I_{\max}} \geq 0.7]$\;
\Return $\langle S, \phi \rangle$
\caption{Rehearse-then-Present}
\label{alg:rehearse}
\end{algorithm}

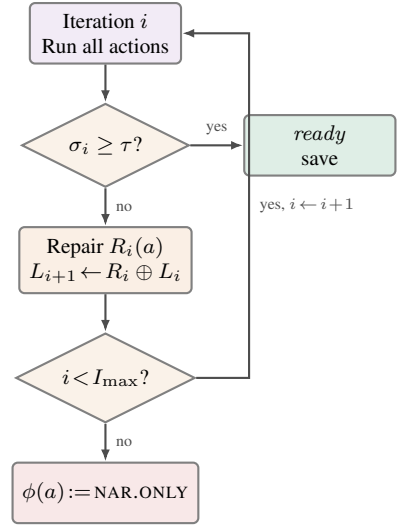
\begin{figure}[t]
\centering
\begin{tikzpicture}[
  font=\scriptsize,
  every node/.style={align=center},
  box/.style={rectangle,rounded corners=2pt,draw=black!50,thick,minimum height=0.8cm,minimum width=2.0cm,fill=white},
  decision/.style={diamond,aspect=2,draw=black!50,thick,minimum width=2.2cm,minimum height=0.5cm,fill=rhorange!10},
  arr/.style={-{Latex[length=1.6mm]},thick,black!70},
]
\node[box,fill=rhpurple!10] (start) {Iteration $i$\\Run all actions};
\node[decision,below=0.5cm of start] (check) {$\sigma_i \geq \tau$?};
\node[box,fill=rhgreen!15,right=0.7cm of check] (ready) {$\mathit{ready}$\\save};
\node[box,fill=rhorange!12,below=0.5cm of check] (repair) {Repair $R_i(a)$\\$L_{i+1}\!\leftarrow\!R_i \oplus L_i$};
\node[decision,below=0.5cm of repair] (iter) {$i\!<\!I_{\max}$?};
\node[box,fill=rhred!12,below=0.5cm of iter] (degrade) {$\phi(a)\!:=\!\textsc{nar.only}$};

\draw[arr] (start) -- (check);
\draw[arr] (check) -- node[above,font=\tiny]{yes} (ready);
\draw[arr] (check) -- node[right,font=\tiny]{no} (repair);
\draw[arr] (repair) -- (iter);
\draw[arr] (iter.east) -- ++(0.7,0) |- node[pos=0.25,right,font=\tiny]{yes, $i\!\leftarrow\!i\!+\!1$} (start.east);
\draw[arr] (iter) -- node[right,font=\tiny]{no} (degrade);
\end{tikzpicture}
\caption{Rehearsal repair loop. Convergence at $\sigma_i \geq \tau$; otherwise, repair and iterate up to $I_{\max}$, then degrade residual failures to \textsc{narration\_only}.}
\label{fig:rehearse}
\end{figure}

\section{Phase 5: Live Presentation Runtime}
\label{sec:runtime}

The runtime serves the rehearsed demo as an interactive experience. Server-side state is held by a per-session \emph{DemoSession} containing $\langle S^*, \Phi^*, K^*, U_\mathrm{app}\rangle$ and a \emph{DemoDriver} state machine; a reverse proxy serves the original target application under a same-origin path so it can be embedded in an iframe and dispatched against from the browser DOM with the same locator strategies the rehearser validated.

\subsection{Synchronization invariant}
\label{sec:sync}

For each segment $\sigma = \langle \mathit{text}, a \rangle$ with $a \neq \emptyset$, let $t_{\mathrm{audio\_end}}^{(\sigma)}$ denote the wall-clock time at which the segment's narration audio finishes, and $t_{\mathrm{action}}^{(\sigma)}$ the wall-clock time at which the action is dispatched in the iframe. The runtime preserves:

\begin{invariant}[Segment synchronization]
\label{inv:sync}
For every segment $\sigma$ with non-null action, $t_{\mathrm{action}}^{(\sigma)} = t_{\mathrm{audio\_end}}^{(\sigma)}$.
\end{invariant}

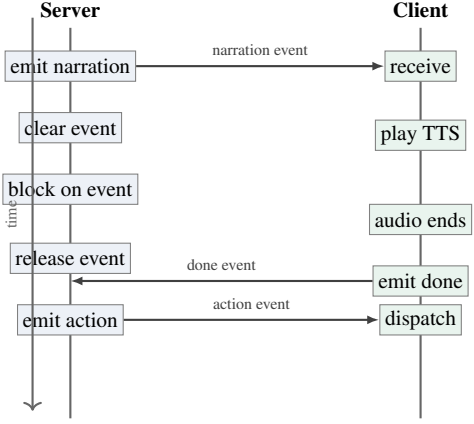
\begin{figure}[t]
\centering
\begin{tikzpicture}[
  font=\scriptsize,
  every node/.style={align=center},
  lifeline/.style={thick,black!60},
  msg/.style={-{Latex[length=1.6mm]},thick,black!75},
  block/.style={rectangle,draw=black!50,fill=rhblue!8,minimum width=0.5cm,minimum height=0.4cm,inner sep=2pt},
  blockc/.style={rectangle,draw=black!50,fill=rhgreen!10,minimum width=0.5cm,minimum height=0.4cm,inner sep=2pt},
]
\node (sl) {\textbf{Server}};
\node[right=3.6cm of sl] (cl) {\textbf{Client}};
\draw[lifeline] (sl) -- ++(0,-5.4);
\draw[lifeline] (cl) -- ++(0,-5.4);

\node[block,below=0.3cm of sl] (s1) {emit narration};
\node[blockc] at (cl|-s1) (c1) {receive};
\draw[msg] (s1) -- node[midway,above,font=\tiny]{narration event} (c1);

\node[block,below=0.4cm of s1] (s2) {clear event};

\node[blockc,below=0.5cm of c1] (c2) {play TTS};

\node[block,below=0.4cm of s2] (s3) {block on event};

\node[blockc,below=0.7cm of c2] (c3) {audio ends};
\node[blockc,below=0.4cm of c3] (c4) {emit done};
\draw[msg] (c4) -- node[midway,above,font=\tiny]{done event} (s3 |- c4);

\node[block,below=0.5cm of s3] (s4) {release event};
\node[block,below=0.4cm of s4] (s5) {emit action};
\node[blockc] at (cl|-s5) (c5) {dispatch};
\draw[msg] (s5) -- node[midway,above,font=\tiny]{action event} (c5);

\draw[->,thick,black!60] (-0.5,-0.1) -- ++(0,-5.2) node[midway,font=\tiny,rotate=90,anchor=south,yshift=2pt]{time};
\end{tikzpicture}
\caption{Segment-completion handshake. The server emits a narration event, blocks until the client plays the TTS audio to completion and signals back, and only then emits the action event. The protocol encodes \Cref{inv:sync}: action dispatch is anchored to the audio-end signal rather than to a predicted timestamp.}
\label{fig:sync}
\end{figure}

The invariant is maintained by a \emph{segment-completion handshake} (Figure~\ref{fig:sync}). The server emits a narration event and clears an internal $\mathrm{narration\_event}$. The client begins TTS playback while the server blocks on $\mathrm{narration\_event}.\mathrm{wait}(\Delta_{\max})$. The client's audio \texttt{onended} fires a \texttt{narration\_done} message that releases the wait, and the server then emits the action event. Provided $\Delta_{\max}$ exceeds the maximum permitted audio length, the invariant holds across variable TTS latency, network jitter, and prosody-induced length variation, yielding zero word-offset drift across an arbitrary number of segments.

\subsection{Three-tier TTS fallback}
\label{sec:tts}

The narration audio path uses a three-tier fallback chain to preserve Invariant~\ref{inv:sync} under provider variability. Let $\mathcal{S}_1$ denote streaming TTS over an OpenAI-compatible \texttt{/audio/speech} endpoint, $\mathcal{S}_2$ non-streaming TTS over the same endpoint, and $\mathcal{S}_3$ the browser-native \texttt{speechSynthesis}. The runtime selects the lowest-index source that emits an \texttt{onended} event; the handshake protocol is identical across all three. This decouples the synchronization argument from any specific TTS provider.

\subsection{Real-time voice question answering}
\label{sec:voice}

User speech is bridged from the browser microphone to a server-VAD speech-to-speech endpoint via a WebSocket relay. The client opens a control channel to the server; the server retrieves a knowledge slice $K_q^* \subseteq K^*$ of bounded length and establishes a second WebSocket to the speech-to-speech endpoint with a session configured for PCM16 input and output, both text and audio modalities, server-side voice activity detection with a fixed threshold and silence-duration parameter, an input transcription model, a configured voice, and an instruction string $I = I_0 \,\Vert\, K_q^*$ where $I_0$ is a generic role preamble. Browser-captured audio frames are forwarded as appended buffer events; the streamed audio response is relayed back to the browser for playback.

\paragraph{Turn latency.}
Per audience turn,
\begin{equation}
T_{\mathrm{turn}} \;=\; T_{\mathrm{up}} + T_{\mathrm{vad}} + T_{\mathrm{stream}} + T_{\mathrm{down}},
\label{eq:turnlat}
\end{equation}
where $T_{\mathrm{up}}$ is end-of-speech to server VAD acceptance, $T_{\mathrm{vad}}$ the configured silence padding, $T_{\mathrm{stream}}$ the LLM-to-first-audio latency, and $T_{\mathrm{down}}$ the audio-stream playback start. Compared to a transcribe--think--synthesize REST pipeline, this path eliminates two full network round-trips and one audio re-encode.

\paragraph{Grounding.}
Because $K^*$ is constructed from both $G$ and $\mathcal{C}$, audience questions over user-facing behavior and over architectural detail are answered from the same retrieval substrate. Let $Q$ denote the set of questions issued during a presentation session. We partition $Q$ into a UI-grounded subset $Q_u$ (questions answerable from $G$) and a code-grounded subset $Q_c$ (questions answerable only from $\mathcal{C}$); a strict subset $Q_{u\cap c}$ requires both. The cross-modal merge is a precondition for any grounded answer over $Q_c$ and $Q_{u\cap c}$.

\paragraph{Multilingual interaction.}
The Realtime endpoint accepts arbitrary input languages via the configured input transcription model and produces output in the same language; narration generation is language-agnostic at the LLM interface. The runtime is therefore multilingual at the system level; localization to a specific language requires per-language voice and prompt selection but no architectural change.

\subsection{Iframe and reverse proxy}

The runtime mounts $U_\mathrm{app}$ under the same origin as the demo client via an HTTP reverse proxy that strips \texttt{X-Frame-Options} and CSP headers from proxied responses and rewrites \texttt{Location} redirects to same-host paths. Same-origin embedding admits direct DOM dispatch into the iframe with the locator strategies validated during rehearsal.

\subsection{Recorded demo packages}

The implementation supports an offline package format for asynchronous viewing: per-scene MP3 audio rendered through the same TTS chain, screenshot frames from rehearsal, and JSON metadata aligning frames to narration segments. The recorded path is structurally distinct from the live runtime; it preserves narrative content but loses dialog, which is the property that distinguishes a live demonstration from a video.

\section{Implementation}
\label{sec:impl}

The system is implemented in approximately 6{,}000 lines of Python with a Flask web layer and Playwright for browser control, plus a thin browser-side client. The agent layer is structured under a common contract: each agent extends a base class returning an \emph{AgentResult} with success, typed payload, error, elapsed milliseconds, and a metadata dictionary; per-phase wall-clock latency is captured by a timing wrapper, and a token-usage field is reserved for cost accounting at the LLM-client boundary.

\paragraph{Provider-agnostic LLM client.}
The LLM client is a thin adapter over an OpenAI-compatible HTTP surface (\texttt{/chat/completions}, \texttt{/audio/speech}, \texttt{/audio/transcriptions}, \texttt{/embeddings}) plus a Realtime WebSocket. Endpoint base URL and bearer key are environment-driven; the client supports both standard chat parameters and reasoning-model parameters with automatic switching by model name; a configurable TLS-verify flag accommodates internal gateways. A deployment can route every LLM call to its preferred provider (direct OpenAI, self-hosted inference, or a multi-provider gateway routing to Claude, Gemini, or open-weight models) without modifying any agent. \Cref{tab:provider-bench} reports an in-house benchmark of the same LLM client across six models and three tasks (narration, fast Q\&A, code analysis), three runs per cell, with zero errors across $54$ invocations and an inter-cell mean-latency spread of $1.7$--$7.9$\,s. The same model under two distinct endpoint base URLs (last two rows of \Cref{tab:provider-bench}) yields comparable latencies, supporting the claim that the client runs unmodified across endpoints.

\begin{table}[h]
\centering
\small
\setlength{\tabcolsep}{4pt}
\begin{tabular}{@{}llcc>{\centering\arraybackslash}p{0.1\linewidth}@{}}
\toprule
Model & Endpoint & Narration & Fast Q\&A & Code anal. \\
\midrule
Gemini 2.0 Flash    & Gateway A & 1.81 & 1.69 & 3.04 \\
Gemini 2.5 Flash    & Gateway A & 2.90 & 2.65 & 7.89 \\
Claude Haiku 4.5    & Gateway A & 2.60 & 2.00 & 2.36 \\
Claude Sonnet 4.6   & Gateway A & 3.80 & 3.58 & 4.42 \\
GPT-4.1 Mini        & Gateway A & 2.41 & 2.38 & 3.57 \\
GPT-4.1 Mini        & Direct    & 2.75 & 2.05 & 3.60 \\
\bottomrule
\end{tabular}
\caption{Mean latency (seconds) per LLM call across six models and three tasks, three invocations per cell ($54$ invocations total, zero errors). \emph{Gateway A} is a multi-provider routing service; \emph{Direct} is a direct OpenAI-compatible endpoint. The latency spread across models and tasks motivates the mixed-tier model assignment.}
\label{tab:provider-bench}
\end{table}

\paragraph{Mixed-tier model assignment.}
Reasoning-heavy phases (code analysis, feature merging, deep Q\&A, scripting) default to a reasoning model; latency-sensitive paths (vision summaries, action-loop decisions, locator repair, narration, fast Q\&A) default to a fast multimodal model; speech and embeddings are served by dedicated endpoints. Reasoning-model detection is heuristic over a set of prefixes, automatically switching the request to use \texttt{max\_completion\_tokens} and \texttt{reasoning\_effort} parameters. \Cref{tab:config} lists the default configuration.

\begin{table}[h]
\centering
\small
\setlength{\tabcolsep}{4pt}
\begin{tabular}{@{}llc@{}}
\toprule
Symbol & Parameter & Default \\
\midrule
$\kappa_p$ & Crawl page cap (Phase 1a) & 30 \\
$\kappa_d$ & Crawl depth cap (Phase 1a) & 4 \\
$B$ & Code batch byte budget (Phase 1b) & 60\,KB \\
$|\mathcal{T}|$ & Number of focus tiers & 3 \\
$\rho$ & Demo priority range & $[1, 10]$ \\
$|\mathcal{V}|$ & Action vocabulary size & 7 \\
$|\mathcal{L}|$ & Locator strategies & 6 \\
$\tau$ & Rehearsal convergence threshold & 0.95 \\
$I_{\max}$ & Rehearsal iteration cap & 3 \\
$\tau_{\mathrm{ready}}$ & Degraded-readiness threshold & 0.70 \\
$B_q$ & Q\&A retrieval char budget & 4000 \\
$\Delta_{\max}$ & Narration handshake timeout & 60\,s \\
\midrule
\multicolumn{3}{l}{\textbf{Per-phase model assignment (defaults)}} \\
\midrule
\multicolumn{2}{l}{Code analysis, understanding, scripting, deep Q\&A} & reasoning \\
\multicolumn{2}{l}{Vision summaries, action repair, narration, fast Q\&A} & fast MM \\
\multicolumn{2}{l}{Text-to-speech, transcription, embeddings, realtime} & dedicated \\
\bottomrule
\end{tabular}
\caption{Default configuration of \sys{}. All values are configurable via environment variables; the symbols correspond to the formal definitions in \Cref{sec:prelim,sec:phase4,sec:phase2,sec:runtime}.}
\label{tab:config}
\end{table}

\paragraph{Structured outputs, parallelism, streaming.}
The client exposes JSON-schema-enforced structured output, parallel batch evaluation through a configurable thread pool (default 4 workers, used by the code-batcher of \Cref{sec:code}), and three streaming modes: token streaming over chat completions, chunked TTS streaming, and event streaming for the Realtime session.

\paragraph{Web surface.}
The web layer exposes a setup page with Server-Sent-Events progress streaming over $\langle \mathit{phase}, \mathit{message}, \mathit{percent}\rangle$ records, a demo page hosting the live viewer, and a player page for recorded packages. The REST surface covers scene retrieval, narration (blocking and streaming), text question answering with fast and deep mode selection, text-to-speech (blocking and streaming), transcription, and capability flags. Two WebSocket channels carry stateful flows: a playback channel between the runtime driver and the browser viewer, and a question-answering channel between the browser microphone and the speech-to-speech relay. Pipeline outputs are persisted under a per-project directory keyed by an MD5-derived slug, and a load endpoint restores any saved project as a new session.

\paragraph{Tests.}
The unit-test suite contains 13 files and 129 tests covering the six agents (including success, repair, and degradation paths for the rehearser), the LLM client (including reasoning-model payload construction and JSON-fence stripping), the configuration loader, the GitHub API client (URL parsing for HTTPS / SSH / GitHub Enterprise, tree filtering, blob fetch, retry-on-rate-limit), the auth parser, and the HTTP routes. External APIs and Playwright are mocked at the boundary, allowing the suite to run without network access.

\begin{figure*}[t]
\centering
\begin{tikzpicture}[
  font=\footnotesize,
  every node/.style={align=center},
  comp/.style={rectangle,rounded corners=3pt,draw=black!50,thick,minimum height=0.85cm,minimum width=2.4cm,fill=white},
  cli/.style={comp,fill=rhblue!10},
  srv/.style={comp,fill=rhgreen!10},
  ext/.style={comp,fill=black!5},
  arr/.style={-{Latex[length=2mm]},thick,black!70},
  bidi/.style={{Latex[length=2mm]}-{Latex[length=2mm]},thick,black!70},
]
\node[srv] (driver) {DemoDriver\\(state machine)};
\node[srv,below=0.3cm of driver] (session) {DemoSession\\$\langle S^*, \Phi^*, K^*\rangle$};
\node[srv,below=0.3cm of session] (qa) {QAHandler\\(text + voice)};
\node[srv,right=0.6cm of driver] (proxy) {ReverseProxy};

\node[cli,right=2.2cm of proxy] (viewer) {Browser client};
\node[cli,below=0.3cm of viewer] (iframe) {iframe ($U_\mathrm{app}$)};
\node[cli,below=0.3cm of iframe] (audio) {TTS audio + mic};

\node[ext,above=0.3cm of driver] (tts) {TTS API};
\node[ext,below=0.3cm of qa] (rt) {Realtime API};

\draw[bidi] (driver.north east) to[bend left=18] node[pos=0.5,above=1pt,font=\scriptsize\itshape]{playback} (viewer.north west);
\draw[bidi] (proxy) -- node[above,font=\scriptsize\itshape]{HTTP} (viewer);
\draw[arr] (viewer) -- (iframe);
\draw[arr] (iframe) -- (audio);
\draw[bidi] (driver) -- (tts);
\draw[bidi] (qa) -- (rt);
\draw[bidi] (qa.east) to[bend right=18] node[pos=0.5,below=1pt,font=\scriptsize\itshape]{voice Q\&A} (audio.south west);

\node[draw=black!30,dashed,rounded corners=3pt,fit=(driver)(session)(qa)(proxy),inner sep=6pt,label={[font=\scriptsize\itshape]above left:server}] {};
\node[draw=black!30,dashed,rounded corners=3pt,fit=(viewer)(iframe)(audio),inner sep=6pt,label={[font=\scriptsize\itshape]above right:browser}] {};
\end{tikzpicture}
\caption{Live runtime architecture. The target application is served back to the client under the same origin via the reverse proxy and embedded in an iframe so that the demo client can dispatch actions in its DOM. The driver controls playback over a WebSocket channel; a separate channel relays user speech to a speech-to-speech endpoint with a knowledge slice $K_q^*$ injected as session instructions.}
\label{fig:runtime}
\end{figure*}
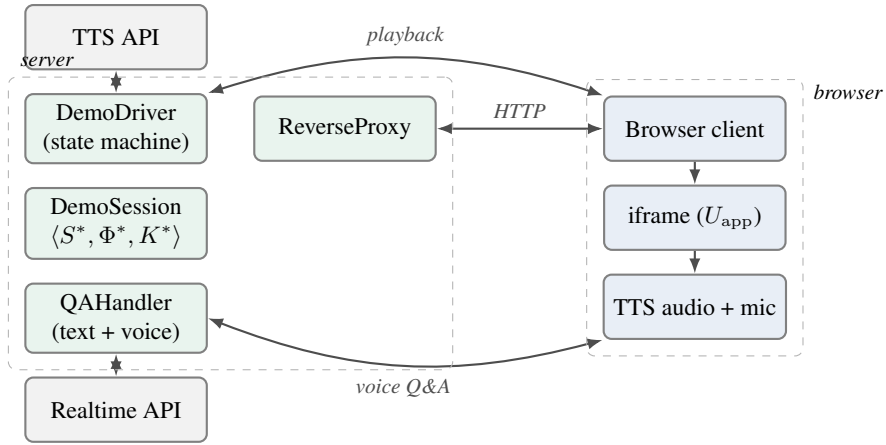

\section{Benchmark Protocol}
\label{sec:eval}

A live demo has more than one thing to be good at, and a single aggregate score over it is misleading. The protocol below specifies ten metrics across four families and a six-category application corpus, and it includes two ablations designed to isolate the contributions of cross-modal merging and of the rehearsal phase. We have not yet run the full protocol; we describe it here so it is reproducible by others, and report what we have measured so far in \Cref{sec:case}.

\subsection{Application corpus}

Let $\mathcal{D}$ be a corpus of open-source web applications stratified across six categories: form-heavy CRUD, dashboards, content management, communications, design tools, and data infrastructure. Table~\ref{tab:corpus} lists a representative cohort. Each application is paired with its source repository; corpus items are selected to span (a) framework family (React/Next, Vue/Nuxt, Django, Rails templates), (b) authentication regime (public, gated with credentials, gated without), (c) interaction richness (form-driven, dashboard, canvas-heavy), and (d) repository size.

\begin{table}[h]
\centering
\small
\begin{tabular}{@{}lll@{}}
\toprule
\textbf{Category} & \textbf{Application} & \textbf{Auth} \\
\midrule
CRUD            & Cal.com         & gated \\
                & Ghost           & gated \\
Dashboard       & Plausible       & gated \\
                & Supabase Studio & gated \\
CMS / records   & NocoDB          & gated \\
                & Directus        & gated \\
Communications  & Mattermost      & gated \\
Design          & Excalidraw      & public \\
                & Penpot          & gated \\
Data platform   & Appwrite        & gated \\
\bottomrule
\end{tabular}
\caption{Representative ten-application corpus stratified across six categories. The corpus targets coverage of authentication regimes and interaction surfaces rather than scale.}
\label{tab:corpus}
\end{table}

\subsection{Pipeline-level metrics}

\begin{definition}[Crawl coverage]
For a discovered navigation graph $G_d$ and the reference set of \emph{user-reachable} pages $G_r$ (manually curated), crawl recall is $\mathrm{CR} = |V(G_d) \cap V(G_r)| / |V(G_r)|$ subject to the crawl budget $\kappa_p$.
\end{definition}

\begin{definition}[Code-extraction recall]
For routes $\hat{\mathcal{R}}$ extracted by Phase 1b and reference routes $\mathcal{R}^*$ enumerated from the framework's routing manifest, route recall is $\mathrm{RR} = |\hat{\mathcal{R}} \cap \mathcal{R}^*| / |\mathcal{R}^*|$.
\end{definition}

\begin{definition}[Feature-extraction precision and recall]
For features $\hat{\mathcal{F}}$ extracted by Phase 2 and a human-annotated reference $\mathcal{F}^*$ keyed by name and entry point, $\mathrm{P}_f = |\hat{\mathcal{F}} \cap \mathcal{F}^*|/|\hat{\mathcal{F}}|$ and $\mathrm{R}_f = |\hat{\mathcal{F}} \cap \mathcal{F}^*|/|\mathcal{F}^*|$, with macro-average $F_1$ over the corpus.
\end{definition}

\subsection{Rehearsal-level metrics}

\begin{definition}[Action success at iteration $i$]
$\mathrm{Succ}_i(S) = \frac{1}{|\mathcal{A}|}\sum_{a \in \mathcal{A}} \mathbf{1}[\phi_i(a) = \textsc{verified}]$.
\end{definition}

\begin{definition}[Repair return on iteration]
$\mathrm{RoI}(i \to i\!+\!1) = \mathrm{Succ}_{i+1}(S) - \mathrm{Succ}_i(S)$.
\end{definition}

\begin{definition}[Degradation rate]
$\mathrm{DR}(S) = \frac{|\{a : \phi(a) = \textsc{narration\_only}\}|}{|\mathcal{A}|}$.
\end{definition}

\subsection{Runtime-level metrics}

\begin{definition}[Live action latency]
The wall-clock time from the segment-completion handshake to the dispatched DOM action, measured client-side, summarized by median and 95th-percentile across all segments in a presentation.
\end{definition}

\begin{definition}[Synchronization drift]
For each segment $\sigma$, $\Delta(\sigma) = t_\mathrm{action}^{(\sigma)} - t_\mathrm{audio\_end}^{(\sigma)}$. Invariant~\ref{inv:sync} predicts $\Delta(\sigma) = 0$ up to network jitter; we report the empirical distribution.
\end{definition}

\begin{definition}[Voice Q\&A turn latency]
$T_\mathrm{turn}$ as decomposed in Eq.~\ref{eq:turnlat}; reported per category and per question class (UI-grounded, code-grounded, or both).
\end{definition}

\begin{definition}[Q\&A grounding accuracy]
For a curated set of questions $Q$ per application with expert-graded reference answers, the proportion of system answers judged \emph{correct and grounded} on a 3-point scale by a blind reviewer, partitioned by $Q_u$, $Q_c$, $Q_{u \cap c}$.
\end{definition}

\subsection{Ablations}
\label{sec:eval-ablations}

The protocol includes two ablations.

\paragraph{A1: Rehearsal on/off.} Compare live $\mathrm{Succ@1}$ when the runtime consumes the unrehearsed Phase 3 script versus the rehearsed Phase 4 script. Hypothesis: rehearsal raises live action success on every category and the gap widens with interaction richness.

\paragraph{A2: Cross-modal vs UI-only.} Compare $\mathrm{P}_f, \mathrm{R}_f$ and the code-grounded Q\&A accuracy on $Q_c \cup Q_{u\cap c}$ between the full pipeline and a variant that drops Phase 1b. Hypothesis: removing code analysis collapses code-grounded Q\&A accuracy and reduces feature-extraction recall on technical features (integrations, data model, scaling), while leaving UI-grounded Q\&A approximately unchanged.

\subsection{Reporting}

For each application, we propose to release the full \emph{site\_map.json}, \emph{script.json}, \emph{rehearsal.json}, and \emph{knowledge.md}, together with the per-iteration success rates, the action-by-action validation log, and the per-question Q\&A grade. Aggregates are reported as macro-averages over $\mathcal{D}$ with 95\% confidence intervals from bootstrap resampling.

\paragraph{Status.} The corpus run over Table~\ref{tab:corpus} has not been completed; we report only the six-session preliminary case study of \Cref{sec:case}. The corpus-level numbers and per-application artifacts will appear in a follow-up revision.

\section{Preliminary Case Study}
\label{sec:case}

Independent of the corpus-level protocol of \Cref{sec:eval}, we report a feasibility case study from running the pipeline end-to-end on four deployed web applications: Application A, an internal enterprise HR/talent-management tool (three sessions of the present implementation, varying crawl parameters); Application B, an internal multi-project synthetic-data-generation research framework with a Flask web dashboard (one session of the present implementation); Application C, an internal AI-platform governance tool (one session of an earlier version of the same multi-agent architecture, retained from saved artifacts); and Application D, the public open-source whiteboard application Excalidraw\footnote{\url{https://excalidraw.com}, source at \url{https://github.com/excalidraw/excalidraw}.} (one session of the present implementation, public-domain reproducible reference point). The case study establishes that the pipeline executes end-to-end across diverse deployed applications including a public-domain target and a canvas-rendered interaction surface, and that the rehearsal repair loop contributes empirically to reaching convergence in at least one observed run. It is not a validation of the individual design choices, which require the protocol of \Cref{sec:eval}; we are explicit about this distinction below.

\subsection{Per-session observations}
\label{sec:case-obs}

\Cref{tab:case-summary} reports the per-application aggregates; per-session detail follows in prose, because the figures are heterogeneous and each requires context.

\begin{table*}[t]
\centering
\small
\setlength{\tabcolsep}{6pt}
\renewcommand{\arraystretch}{1.10}
\begin{tabular}{@{}llccccll@{}}
\toprule
App & Description & Sess.\ & $|\mathcal{A}|$ & $\bar\sigma$ & Iter & Tiers populated & Regime \\
\midrule
\textbf{D} & Excalidraw (public OSS) & \textbf{1} & \textbf{14} & \textbf{1.00} & \textbf{2} & H+S+M & converged via repair \\
\textbf{C} & AI-platform governance & 1 & \textbf{53} & \textbf{0.92} & 3 & H+S+M & near-convergence \\
A & HR/talent-management & 3 & 16--22 ea.\ & 0.31--0.55 & 3 & varies & degraded ($I_{\max}$) \\
B & Synthetic-data dashboard & 1 & 22 & 1.00 & 1 & H only & small-surface easy case \\
\bottomrule
\end{tabular}
\caption{Per-application case-study aggregates over six rehearsal sessions on four deployed applications. $|\mathcal{A}|$: scripted actions. $\bar\sigma$: rehearser internal locator-firing rate (not an external success measure). Iter: rehearsal iterations to terminate. Tiers populated: \textsc{H}=\textsc{hero}, \textsc{S}=\textsc{supp}, \textsc{M}=\textsc{mention}. Regime: which control-flow branch of Algorithm~\ref{alg:rehearse} the run took. Application D (public-domain Excalidraw, convergence reached via the repair step at iteration $2$) and Application C (substantial workload with full tier differentiation, near-convergence) are the strongest evidentiary points; Application A (degraded regime) and Application B (small-surface easy case) are reported in context.}
\label{tab:case-summary}
\end{table*}

We summarize the six sessions in prose below because the per-session figures are heterogeneous and each requires interpretation.

\emph{Application D} (Excalidraw) is the public-domain reference point and the only session in which the rehearsal repair loop is observed to contribute to convergence. The session produced a $7$-scene script ($1$ \textsc{hook} + $4$ \textsc{journey} + $1$ \textsc{hood} + $1$ \textsc{close}) on the canvas-rendered single-page surface, with all three tiers populated ($3$ \textsc{hero} + $3$ \textsc{supp} + $1$ \textsc{mention}) and $14$ scripted actions. Iteration $1$ did not converge at $\tau$; the locator-repair step proposed alternatives and iteration $2$ reached $\bar\sigma = 1.00$ with all $14$ actions verified. Coherent demo structure on a canvas-heavy SPA, full tier differentiation despite a one-URL crawl, and direct evidence that the repair loop drives convergence make this the strongest single session.

\emph{Application C} is the strongest substantial-workload session: a single run on an internal AI-platform governance tool produced an $11$-scene script with $53$ scripted actions and proper four-act distribution ($2$ \textsc{hook} + $7$ \textsc{journey} + $1$ \textsc{hood} + $1$ \textsc{close}) and proper tier differentiation ($6$ \textsc{hero} + $2$ \textsc{supp} + $3$ \textsc{mention}). Of the $53$ actions, $49$ verified and $4$ degraded under the rule of \Cref{sec:phase4}, yielding an internal locator-firing rate $\bar\sigma \approx 0.92$ and a readiness flag of $0.92$. The session is the most substantial workload in the case study and approaches the convergence threshold $\tau = 0.95$ on a non-trivial action count.

\emph{Application A} comprises three sessions of the present implementation on an internal HR/talent-management tool, run with different crawl parameters. Across these sessions the pipeline reached 1, 4, and 29 pages respectively, produced 7, 9, and 11 scenes with 16, 22, and 20 actions, and recorded $\bar\sigma$ values of $0.31$, $0.41$, and $0.55$. All three sessions exited at $I_{\max}$ without converging at $\tau$ and entered the degradation path of \Cref{sec:phase4}. The 1-page session reflects a single-page-application crawl reaching one URL whose elements drove the script; we note this because the page count alone is not an exploration failure when the application is an SPA. The 4-page session produced a knowledge document of $125$ sections, a ratio of about $31$ sections per crawled page that suggests $K^*$ extends beyond what exploration directly grounded; this is an open empirical question that Definition $10$ would answer and is the kind of behavior the protocol is designed to surface.

\emph{Application B} comprises one session of the present implementation on a synthetic-data-generation research dashboard. The crawl reached $2$ pages of the public surface; the script contained $8$ scenes ($1$ \textsc{hook} + $5$ \textsc{journey} + $1$ \textsc{hood} + $1$ \textsc{close}), all tagged \textsc{hero} in the absence of feature differentiation under a small surface; all $22$ scripted actions verified at iteration $1$ ($\bar\sigma = 1.00$), and the repair loop and degradation rule were therefore not exercised in this run. The session is interpreted as the small-public-surface easy case rather than as evidence of reliable convergence.

\subsection{Action latency}

Across the $6$ sessions and $147$ scripted actions, the empirical action-duration distribution is bimodal: short actions (\texttt{wait}, \texttt{scroll}, in-DOM \texttt{click}) complete in milliseconds, while \texttt{navigate} and \texttt{fill} actions on slow pages occupy a long tail extending to the execution timeout. Per-session rehearsal wall-clock time was 10--41\,s for Application A, approximately $24$\,s for Application B, and approximately $36$\,s for Application D over its two iterations, dominated by browser action latency rather than by LLM repair calls.

\subsection{Validated deployment}

In addition to the runs reported above, an earlier integration of the same architecture operates within Application B as an in-tree component of its web layer. The integration provides the same runtime surface as the standalone system---script retrieval, narration, text and voice question answering, text-to-speech, and transcription---together with a real-time speech-to-speech path, and is reachable from the application home page through a top-level navigation link. The deployment serves an authored knowledge document of 1{,}440 lines partitioned into 19 sections, and has been exercised in repeated presentation sessions on Application B in which voice queries were handled through the segment-completion synchronization protocol of \Cref{sec:sync} and the speech-to-speech path of \Cref{sec:voice}. The standalone system described in this paper is the application-agnostic generalization of that integration, sharing the same data structures, the same locator priority $\mathcal{L}$, and the same synchronization invariant.

\subsection{What this case study establishes}

The case study supports four narrow claims. First, the pipeline executes end-to-end without operator intervention on four deployed applications differing in framework, interaction surface, and authentication regime, including a public-domain canvas-rendered reference point (Application D, Excalidraw). Second, on a substantial workload (Application C, $53$ actions, full tier differentiation), an internal locator-firing rate of $\bar\sigma \approx 0.92$ is reached, with the residual handled by the degradation rule. Third, the locator-repair step empirically drives convergence in at least one observed run: Application D fails to converge at iteration $1$, the repair step proposes alternatives, and iteration $2$ reaches $\bar\sigma = 1.00$. Fourth, the wall-clock cost of rehearsal is dominated by browser action latency rather than LLM cost: action durations are bimodal, and per-session totals are 10--41\,s for Application A, approximately $24$\,s for Application B, and approximately $36$\,s for Application D over its two iterations.

\subsection{What this case study does not establish}
\label{sec:case-gaps}

The case study does not validate the individual design contributions; we are explicit about each gap because each maps to a specific item in the protocol of \Cref{sec:eval}.

\paragraph{Repair-loop contribution is partially measured.}
Application D establishes one observed instance in which the repair step drives convergence (iteration $1$ below $\tau$, iteration $2$ at $\bar\sigma = 1.00$). However, the per-iteration $\mathrm{Succ}_i$ trajectories were not retained for the case-study sessions, so the per-iteration return-on-iteration $\mathrm{RoI}(i \to i\!+\!1)$ (Definition 5) is not reported here, and no rehearsal-on-vs-off comparison (Ablation A1) is reported either. Application A's three sessions terminated at $I_{\max}$ without converging, establishing that repair did not reach $\tau$ on those workloads but not separating ``no improvement per iteration'' from ``insufficient iterations.'' Application B converged at iteration $1$ (no repair exercised) and Application C terminated at $I_{\max}$ with $\bar\sigma \approx 0.92$. Quantifying the magnitude of the repair contribution and characterizing its dependence on application complexity requires the runs specified in \Cref{sec:eval-ablations}.

\paragraph{Cross-modal contribution is unmeasured.}
Ablation A2 (full pipeline vs.\ Phase 1b removed) is the experiment that would isolate the contribution of source-code analysis to feature extraction and to code-grounded Q\&A accuracy. We have not run it. The case study therefore does not justify the cross-modal merge over a UI-only baseline.

\paragraph{Output quality is unmeasured.}
$\bar\sigma$ measures whether the rehearser's locators fired; it does not measure whether the script presents the right features in the right order with appropriate narration. Feature-extraction precision and recall (Definition 3) require a human reference $\mathcal{F}^*$ which we have not yet annotated. Demo coherence and coverage are not measured. A session with low $\bar\sigma$ on a well-covered crawl and a session with high $\bar\sigma$ on a one-page crawl can in principle be ranked in opposite directions by the metric reported here and by a downstream quality measure; the case study cannot distinguish these.

\paragraph{Q\&A grounding accuracy is unmeasured.}
The voice and text Q\&A paths retrieve from the generated knowledge document $K^*$ but the case study does not measure whether retrieved answers are factually grounded in $\mathcal{F}$ and $\mathcal{C}$. Definition 10 specifies the protocol; we have not curated the question set $Q$ or rated answers. The high section count for Run A.3 (125 sections from 4 crawled pages) is a flag for this future work: it suggests $K^*$ may extend beyond what exploration directly grounded, which is precisely what Definition 10 would catch.

\paragraph{Convergent-regime behavior is observed only on two sessions.}
Application B reaches $\bar\sigma = 1.00$ on a small public surface and Application C reaches $\bar\sigma \approx 0.92$ on a substantial surface. These are encouraging single-session observations, not evidence that convergence is reliable across applications; the protocol of \Cref{sec:eval} is required to characterize the convergence rate and its dependence on application complexity over the corpus.

\paragraph{Summary.}
The case study demonstrates that the pipeline runs and that the artifacts described in \Cref{sec:prelim,sec:phase1,sec:phase2,sec:phase3,sec:phase4} are produced. Whether the produced artifacts are good demonstrations, whether each design choice contributes positively, and whether the reported $\bar\sigma$ predicts downstream quality remain open and require the protocol of \Cref{sec:eval}. We commit to running and reporting that protocol in subsequent revisions.

\section{Limitations}
\label{sec:limits}

\paragraph{Crawl budget.} The default $\kappa_p = 30$, $\kappa_d = 4$ are calibrated for representative web-application surfaces. Deep enterprise applications with many gated workflows are systematically under-explored without per-application tuning.

\paragraph{Authentication.} Vision-assisted login resolves unauthenticated public-area exploration. Authenticated demos require an injected session cookie, scripted credentials, or a bearer token; each requires operator setup.

\paragraph{Canvas-rendered surfaces.} Applications whose primary interaction surface is HTML5 canvas (diagram editors, design tools) are difficult: the DOM does not expose interactive elements and the locator priority order in $\mathcal{L}$ is uninformative. Vision-only locators are a separate research direction.

\paragraph{Repair under partial observability.} The current locator-repair LLM call passes target metadata and an error message but not the post-failure DOM. Conditioning the repair prompt on a structured DOM extract is a low-cost extension expected to materially improve repair RoI.

\paragraph{Single-shot understanding.} Phase 2 is a single non-iterative LLM call without schema-validation re-prompt. Malformed or under-specified outputs degrade gracefully into missing demo content rather than surfacing as explicit errors.

\paragraph{Multilingual surface area.} The voice Q\&A path is multilingual end-to-end and narration generation is language-agnostic at the model interface. The remaining English-specific surfaces---the REST transcription path's default language and the default narration prompt---are reconfigurable per request, but a fully localized presentation surface (per-language voice presets, locale-specific narrative patterns) is a deployment task.

\paragraph{Generalization.} The pipeline assumes a server-rendered or SPA web application accessible at a URL with a parallel public source repository. Mobile, native desktop, and closed-source SaaS targets are out of scope; each is a separate adaptation of the same multi-phase pattern.

\paragraph{Safety, privacy, and consent.} A demo agent that crawls a target application and reads its source code raises consent and exfiltration concerns analogous to those discussed in~\cite{vardanyan2025browser}. The system is intended for first-party use on the operator's own product; a hosted multi-tenant deployment requires an additional trust layer that we do not address.

\section{Discussion}
\label{sec:disc}

\paragraph{Rehearse-then-present as a design pattern.}
The architectural contribution of Phase 4 is the relocation of the agent's failure surface from runtime to a controlled offline phase. The runtime consumes a validated trace, supplemented by an explicit degradation fallback for any action the rehearsal loop could not validate. The pattern generalizes beyond demonstrations to any browser-agent setting in which the marginal cost of a runtime failure exceeds the marginal cost of an offline retry: scheduled customer walk-throughs, training-environment scripted lessons, and reviewer-facing executive demonstrations are immediate examples. No new learning algorithms are required for the pattern; the implementation cost reduces to a pre-flight execution phase and a typed degradation rule.

\paragraph{Segment as the synchronization unit.}
Word-offset synchronization fails under variable TTS latency and under prosody-driven departures from the underlying text. The segment-completion invariant of \Cref{sec:sync} replaces a continuous timestamp-estimation problem with a discrete server--client handshake whose correctness follows from the unit of authoring rather than from real-time signal estimation. The invariant is independent of the specific TTS provider (\Cref{sec:tts}) and applies to any system in which spoken narration is paired with discrete actions on a controllable surface.

\paragraph{Cross-modal grounding for narrative planning.}
Prior work that combines code analysis with UI signals does so to inform task-time execution. \sys{} uses the same combination upstream, at planning time, to determine which features are demonstrated, what architectural content is foregrounded in Act 3, and which subset $Q_c \cup Q_{u \cap c}$ of audience questions is answerable from the merged knowledge document. The contribution is the placement of the merge in the planning phase rather than in the execution phase.

\paragraph{Knowledge document as a shared retrieval substrate.}
Both narration generation (\Cref{sec:phase3}) and runtime question answering (\Crefrange{sec:voice}{sec:runtime}) retrieve from the same Markdown document $K^*$. This sharing is the structural reason that the runtime's spoken responses are consistent with the script: the agent answers from the same representation it presents from. The voice path differentiates the runtime from a video player by converting the user-side interaction model from passive playback to grounded interruption.

\section{Conclusion}

\sys{} is an end-to-end system for rehearsed live demonstrations of web applications. The pipeline executes UI exploration in parallel with source-code analysis, merges both signals into features tagged with discrete focus tiers, generates a four-act narrative whose actions reference observed UI elements, and validates each action in a real headless browser before deployment. At runtime, the rehearsed script is served through a same-origin reverse proxy embedded in the user's browser; narration audio is synchronized to browser actions through a per-segment handshake; and user speech is routed to a server-VAD speech-to-speech endpoint grounded in the generated knowledge document. The paper formalizes four design choices---cross-modal feature representation, grounded scripting with semantic locators, rehearse-then-present with explicit convergence and degradation, and a segment-completion synchronization invariant---and proposes a benchmark protocol comprising ten metrics across six application categories. A six-session case study spanning four deployed applications exercises multiple rehearsal regimes and supports the claim that the design preserves runtime completeness across the convergence threshold.

\section*{Reproducibility}
The implementation, configuration, and prompts are organized to be reproducible from \texttt{pip install -e .} plus \texttt{playwright install chromium}. Default models are configurable via environment variables; the system runs unmodified against any OpenAI-compatible API endpoint, including local inference servers and multi-provider gateways.

\bibliographystyle{plain}

\end{document}